\documentclass{ifacconf}

\usepackage[T1]{fontenc}
\usepackage{graphicx}      
\usepackage{natbib}        
\usepackage{theorem}
\usepackage{amsmath,amssymb,amsfonts,amsbsy}
\usepackage{dsfont}
\usepackage{xcolor}
\usepackage{comment}
\usepackage{svg}
\usepackage{algorithm}
\usepackage{algpseudocode}

\newcommand{\setR}{\mathbb{R}}
\newcommand{\semposR}{\mathbb{R}_{\geq0}}
\newcommand{\posR}{\mathbb{R}_{>0}}

\newcommand{\mask}{\mathfrak{m}}

\newcommand{\ones}{\mathds{1}}

\definecolor{Arbitrary}{HTML}{FF891D}
\definecolor{PageRank}{HTML}{B51AFA}

\begin{document}
\begin{frontmatter}

\title{The impact of sensor placement on graph-neural-network-based leakage detection
\thanksref{footnoteinfo}} 

\thanks[footnoteinfo]{This work was performed in the cooperation framework of Wetsus, European Centre of Excellence for Sustainable Water Technology (www.wetsus.nl). Wetsus is co-funded by the European Union (Horizon Europe, LIFE, Interreg and EDRF), the Province of Fryslân and the Dutch Government: Ministry of Economic Affairs (TTT, SBO $\&$ PPS-I/TKI Water Technology), Ministry of Education, Culture and Science (TTT $\&$ SBO) and Ministry of Infrastructure and Water Management (National Growth Fund - UPPWATER). The authors like to thank the participants of the research theme “Smart Water Grids” for the fruitful discussions and their financial support.}

\author[First]{J.J.H.v.Gemert} 
\author[First]{V. Breschi} 
\author[Second]{D.R. Yntema}
\author[Third]{K.J. Keesman}
\author[First]{M. Lazar}

\address[First]{Control Systems Group, Dept. of Electrical Engineering, Eindhoven University of Technology, 5600MB Eindhoven, The Netherlands (e-mail: J.J.H.v.Gemert@tue.nl).}
\address[Second]{Wetsus, Centre of Excellence for Sustainable Water Technology, 8900MA Leeuwarden, The Netherlands.}
\address[Third]{Mathematical and Statistical Methods – Biometris, Wageningen University, 6708PB Wageningen, the Netherlands}

\begin{abstract}                
Sensor placement for leakage detection in water distribution networks is an important and practical challenge for water utilities. Recent work has shown that graph neural networks can estimate and predict pressures and detect leaks, but their performance strongly depends on the available sensor measurements and configurations. In this paper, we investigate how sensor placement influences the performance of GNN-based leakage detection. We propose a novel PageRank-Centrality-based sensor placement method and demonstrate that it substantially impacts reconstruction, prediction, and leakage detection on the EPANET Net1.
\end{abstract}


\end{frontmatter}

\section{Introduction}\label{Sec: Introduction}
Leakages in water distribution networks (WDNs) can lead to considerable water losses, increased operational costs, and infrastructure damage. Beyond these financial and structural impacts, leakages may also pose public-health risks due to the potential ingress of contaminants \citep{XU2014955}. These challenges are becoming even more pressing as water scarcity becomes an increasing problem under the effects of climate change, highlighting the need for effective leakage detection and leakage reduction strategies \citep{annaswamy2024control,Leakage_development}.
In this context, effective leakage detection is a critical concern for many water utilities, which typically rely on two established methods. The first is Minimum Night Flow (MNF) analysis, which assumes that night-time demand is low and predictable, thus any significant increase in flow during these hours can be interpreted as a leakage \citep{al2018modelling,lee2022prediction}. 
However, such assumptions often do not hold in practice due to irregular demand patterns, which can limit the reliability of MNF-based detection. The second approach is model-based residual analysis, in which measured and reconstructed pressures are compared with predicted pressures obtained from hydraulic models such as EPANET or InfoWorks \citep{hu2021review}. A leak is indicated when the resulting residuals, i.e., the difference between reconstructed and predicted pressures, exceed a predefined threshold. The reliability of this method depends strongly on model quality, which in turn relies on accurate real-time data, demand estimates, and hydraulic parameter values. Thus, uncertainties in any of these key factors can significantly influence leakage detection performance.

These challenges have motivated interest in learning-based methods that reconstruct pressures directly from sensor measurements and do not depend on accurate hydraulic parameters. More specifically, graph neural networks (GNNs) have gained attention, as they exploit the WDN’s graph topology to learn how pressure propagates through the network, enabling reconstruction at unobserved junctions even with only a few sensors \citep{hajgato2021reconstructing}.
Several GNN architectures have been explored in this context, including spectral methods such as ChebNet \citep{hajgato2021reconstructing} and hierarchical graph networks such as the graph U-Net in \citep{truong2024graph}.
Neural networks have also been applied directly to leakage detection. Examples include the ANN-based pipeline leakage diagnosis in \citep{Perez2021ANN}, which detects leak-induced pressure disturbances from sensor data, and the GNN reconstructor–predictor framework in \citep{gardharsson2022graph}, which identifies leaks by comparing reconstructed and one-step-ahead predicted pressures.
Despite these advances, the effectiveness of such GNN-based methods still depends on a fundamental factor: the underlying sensor configuration of the network. In most existing works, the sensor layout is fixed a priori or chosen randomly \citep{gardharsson2022graph}, and its influence on reconstruction accuracy, prediction performance, or the overall leakage-detection process is not explicitly examined.
However, this part is critical, as the informativeness of the pressure measurements, and thereby the overall performance, is intrinsically tied to the sensor configuration.

Sensor placement for leakage detection has been widely studied in both industry and academia. In practice, utilities often rely on simulation-based methods, where different sensor configurations are evaluated against a set of explored simulated leak scenarios \citep{santos2022pressure,ROMEROBEN202254}. These approaches target leakage detection and reflect realistic operating conditions. 
However, their performance is limited to the finite set of simulated scenarios, and they do not provide formal guarantees.
To address this, recent works have explored observability theory as a basis for sensor placement. In this framework, ensuring observability guarantees that pressures at unmeasured junctions can be reconstructed from the available sensor data over a finite time. For example, approaches such as \citep{geelen2021optimal,bopardikar2021randomized} aim to maximize leak detectability by maximizing the ``degree'' of observability. These methods, however, require accurate hydraulic parameters, demand patterns, and full model knowledge, which can be difficult to obtain in practice.
To overcome this challenge, recent work has proposed structural observability–based sensor placement methods that rely solely on the WDN topology and guarantee observability without requiring any hydraulic information \citep{VanGemert2024ECC,vangemert2025scalablesensorplacementcyclic}. While effective, the resulting sensor configurations are often conservative and may demand more sensors than is feasible in practice given installation and cost constraints.

Given the above challenges, this paper proposes a novel, scalable, topology-based, and model-free sensor placement method based on PageRank Centrality. We examine how the resulting PageRank-Centrality-based sensor placement fits with the ChebNet reconstructor–predictor framework and how it affects pressure reconstruction, one-step prediction, and residual-based leakage detection. Using the EPANET Net1 benchmark, we compare this approach with an arbitrary configuration and evaluate the influence of sensor placement on reconstruction accuracy, prediction performance, and leakage detection.

The remainder of this paper is structured as follows.  
Section \ref{Sec: Preliminaries and Problem Statement} introduces the necessary preliminaries, including basic graph-theoretic concepts, the WDN setting, and the problem formulation.  
Section \ref{Sec: Main results} presents the main contributions, starting with the PageRank-Centrality-based sensor placement, followed by the ChebNet reconstructor–predictor framework for pressure reconstruction and one-step prediction, and the associated residual-based leakage detection mechanism. 
Section \ref{Sec: Simulation results} evaluates the approach on the EPANET Net1 benchmark.  
Finally, Section \ref{Sec: Conclusion and Future Work} summarizes the findings and outlines directions for future research.

\begin{figure}[b]
    \centering
\vspace{-0.15cm}\includegraphics[width=0.6\linewidth]{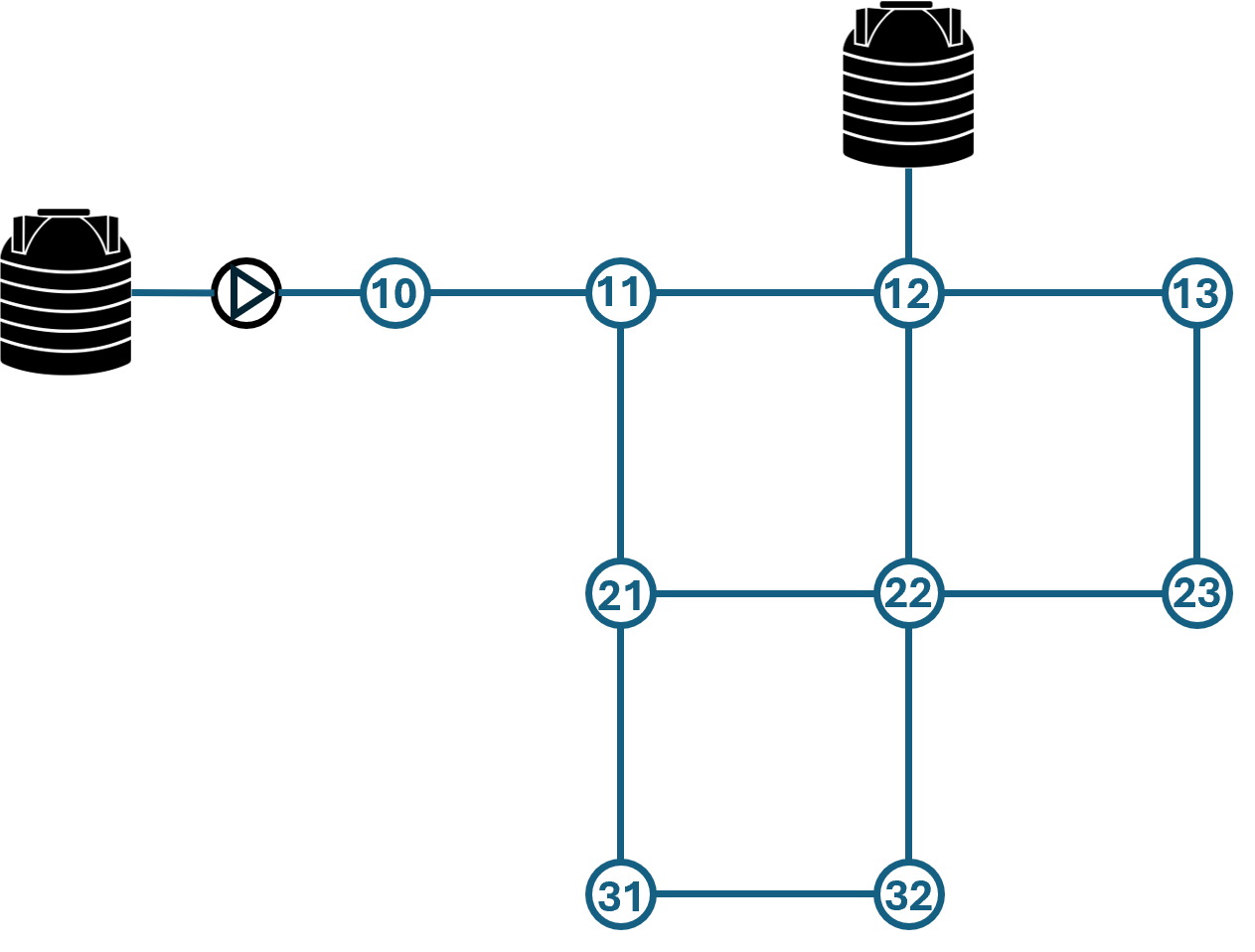}
    \vspace{-0.25cm}\caption{Illustration of EPANET Net1 WDN consisting of two
reservoirs/tanks (black icons), one pump (black circle), and
junctions (blue circles) connected through pipes (blue lines).}
    \label{fig:wdn_example}
\end{figure}

\emph{Basic notation:}
Let $\setR$ denote the field of real numbers, and let $\semposR$ and $\posR$ denote the sets of non-negative and positive reals, respectively.
For a vector $x \in \setR^{n}$, $x_i$ denotes its $i$-th element, and $\ones_n \in \setR^{n}$ denotes the vector of all ones. 
For a matrix $A \in \setR^{n \times n}$, $A^{-1}$ denotes its inverse, and $a_{ij}$ the element on row $i$ and column $j$. The $n \times n$ identity matrix is denoted by $I_n$.

\begin{figure*}[t]
    \centering
    \includegraphics[scale=0.5]{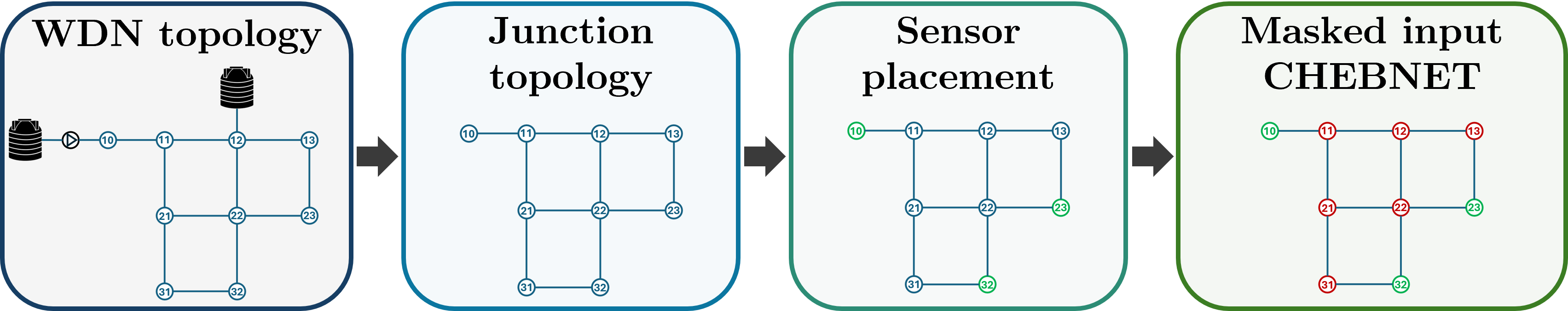}
    \caption{Processing pipeline from WDN topology to Chebnet input.}
    \label{fig:Sensor_flow}
\end{figure*}

\section{Preliminaries and Problem Statement}\label{Sec: Preliminaries and Problem Statement}
In this section, we introduce the basic graph-theoretic concepts that provide the foundations for the sensor placement algorithm, GNNs, and WDN modeling, along with the general WDN setting and the problem formulation.

\subsection{Basic graph notions $\&$ WDNs}
In this paper, we consider only undirected and unweighted graphs. A graph is undirected if every edge is bidirectional, i.e., an edge from node $i$ to node $j$ implies an edge from $j$ to $i$, and it is unweighted if all edge weights are equal to $1$.
We define a graph as $\mathcal{G} = (\mathcal{V},\mathcal{E})$, 
where $\mathcal{V}=\{v_i\}_{i=1}^{n}$ is the set of nodes, and
$\mathcal{E}=\{e_k\}_{k=1}^{m}$ is the set of edges. 
The adjacency matrix $A \in \mathbb{R}^{n\times n}$ represents the topology of such a graph, where
\begin{equation}\label{eq: Adjacency}
a_{ij} =
\begin{cases}
1, & \text{if edge } e_k \text{ connects } v_i \text{ and } v_j,\\[2pt]
0,   & \text{otherwise}.
\end{cases}
\end{equation}
Given the adjacency matrix, the \emph{out-degree} and \emph{in-degree} of a node $v_i$ are defined as
\begin{equation*}
d_i^{\mathrm{out}} = \sum_{j=1}^n a_{ij},
\qquad
d_i^{\mathrm{in}}  = \sum_{j=1}^n a_{ji}.
\end{equation*}
For an undirected graph, we have that $a_{ij} = a_{ji}$ for all node pairs $(i,j)$. As a consequence, the in-degree and out-degree of each node coincide, and we simply write
\begin{equation*}
d_i^{\mathrm{in}} =d_i^{\mathrm{out}} =d_i = \sum_{j=1}^{n} a_{ij},
\end{equation*}
whose values form the diagonal entries of the degree matrix
\begin{equation}\label{eq: DegreeMatrix}
D = \mathrm{diag}(d_1,\dots,d_n).
\end{equation}
The graph Laplacian is then defined as
\begin{equation}\label{eq: Laplacian}
L = D - A,
\end{equation}
and captures how each node relates to its neighbors and its connectivity. 
Because of this structural role, the Laplacian is central to spectral GNNs and forms the foundation of the Chebyshev graph convolutions used later in this paper.

Before formulating the problem, we first describe the WDN setting in which it is posed. A WDN is an infrastructure system that delivers pressurized drinking water from supply sources to consumers through an interconnected pipe network. A WDN typically consists of reservoirs and tanks that provide hydraulic head, pumps that maintain pressure, pipes that transport water, and junctions where water is distributed towards consumer demands. An illustration of a small WDN is shown in Fig. \ref{fig:wdn_example}. The topology of this network can be represented by an adjacency matrix of the form defined in \eqref{eq: Adjacency}, which captures how these junctions, tanks, and reservoirs are linked through the pipe network.

\subsection{Problem formulation}
Given a WDN such as the one illustrated in Fig. \ref{fig:wdn_example}, we first establish the setting for the sensor placement problem. In practice, pressure sensors are typically installed only at junctions, whereas tanks, reservoirs, and pumps are often already equipped with sensors and treated as boundary nodes within the hydraulic model. Consistent with this, we therefore consider only junctions as candidate locations for pressure sensors. The overall workflow of the sensor placement pipeline is shown in Fig. \ref{fig:Sensor_flow}.

Building on this setting and the challenges highlighted in Section \ref{Sec: Introduction}, we investigate the following question:
\emph{How can we exploit graph topology and connectivity for sensor placements that enhance pressure reconstruction and prediction accuracy, thereby improving leakage detection in GNN-based methods?}

The main problem will be addressed in three steps:
(i) a topology-based sensor placement method is introduced, based on PageRank Centrality; 
(ii) a ChebNet reconstructor–predictor framework is used for pressure reconstruction and one-step prediction from the resulting sparse measurements; and
(iii) residual analysis is applied to detect leakages by comparing reconstructed and predicted pressures.  
These components are then connected through a consistent graph construction that ensures the PageRank-Centrality-based sensor placement and the ChebNet architecture operate on the same underlying topology.

\section{Main results}\label{Sec: Main results}
In this section, we develop the methodology outlined above. 
We first introduce a topology-based sensor placement method using PageRank Centrality, which selects an informative sparse set of sensors. 
We then describe the ChebNet architecture used for pressure reconstruction and one-step prediction \citep{hajgato2021reconstructing}, together with the residual-based leakage detection framework from \citep{gardharsson2022graph}. 
Finally, we demonstrate how the PageRank-based sensor placement aligns with the ChebNet architecture through a consistent graph construction.

\subsection{PageRank-Centrality-based Sensor Placement}
\label{Subsec: PageRankSensorPlacement}
Using the graph representation introduced in Section \ref{Sec: Preliminaries and Problem Statement}, we now define a topology-based, model-free sensor placement method. The approach is based on PageRank Centrality \citep{brin1998anatomy}, which evaluates how a simple random walk moves through the network. Junctions that the random walk reaches least often are particularly interesting to measure when only a small number of sensors can be installed. Their pressure values are the most difficult for a GNN to estimate from neighboring junctions.
To formalize this method, we express PageRank Centrality on the WDN topology. We begin with the random-walk transition matrix
\begin{equation}\label{eq:PR_transition}
    M = A D^{-1},
\end{equation}
where  $A $ is the adjacency matrix \eqref{eq: Adjacency} and  $D $ the degree matrix \eqref{eq: DegreeMatrix}.  
The entry  $M_{ji} $ gives the probability that a random walker located at junction  $i $ moves to junction  $j $ in one step.
The PageRank algorithm augments this basic random walk using a uniform vector
\begin{equation}
        v = \tfrac{1}{n}\mathbf{1}^n.
\end{equation}
and a damping factor  $\alpha \in (0,1) $, resulting in the iteration
\begin{equation}\label{eq: PageRank}
    p^{(k+1)} = \alpha M p^{(k)} + (1-\alpha)v,
\end{equation}
starting from  $p^{(0)} = v $.
At each iteration, the product  $Mp^{(k)} $ performs a random-walk update, distributing the current scores across neighboring junctions according to  $M $, while the term  $(1-\alpha)v $ redistributes a uniform amount across all junctions.  
Repeated application of \eqref{eq: PageRank} converges to a unique vector  $p^\ast $ satisfying
\begin{equation}\label{eq:PR_fixedpoint}
    p^\ast = (I - \alpha M)^{-1}(1-\alpha)v,
\end{equation}
which quantifies how accessible each junction is under the modified random walk: frequently visited junctions obtain larger values, whereas rarely visited ones receive smaller values.

Algorithm \ref{Alg:PageRankSensorPlacement} demonstrates the resulting sensor placement using this PageRank Centrality formulation. 
For a desired number of sensors  $s $, the algorithm constructs the degree and transition matrices (line 1), applies the PageRank iteration until convergence (lines 2–5), and then selects the  $s $ junctions with the smallest PageRank values (line 6) for the chosen sensor locations. These junctions provide an informative sensor location, as their pressures are difficult to estimate from neighboring measurements based on their PageRank Centrality value.

\begin{algorithm}[t]
\caption{PageRank-Centrality-based Sensor Placement}
\label{Alg:PageRankSensorPlacement}
\begin{algorithmic}[1]
\Require Adjacency matrix $A \in \{0,1\}^{n \times n}$, 
number of sensors $s$, $\alpha \in (0,1)$, tolerance $\varepsilon>0$.
\Ensure  Sensor index set $\mathcal{S}$.
\State $d \gets A\mathbf{1}$, \quad $D \gets \mathrm{diag}(d)$, \quad $M \gets AD^{-1}$
\State $p^{(0)} \gets \tfrac{1}{n}\mathbf{1}$, \quad $v \gets \tfrac{1}{n}\mathbf{1}^n$
\Repeat
    \State $p^{(k+1)} \gets \alpha M p^{(k)} + (1-\alpha)v$
\Until $\|p^{(k+1)} - p^{(k)}\|_2 \le \varepsilon$
\State Sort $p^{(k+1)}$ in ascending order and let $\mathcal{S}$ be the indices of the $s$ smallest entries
\Return $\mathcal{S}$
\end{algorithmic}
\end{algorithm}

Having defined this topology-driven sensor placement method, we now introduce the ChebNet architecture, which uses the resulting sparse pressure measurements to estimate the pressures throughout the WDN.

\subsection{ChebNet Architecture for Pressure Reconstruction and Prediction}\label{Subsec: ChebNet Architecture for Pressure Reconstruction and Prediction}
\begin{figure*}[t]
    \centering
    \includegraphics[width=0.95\linewidth]{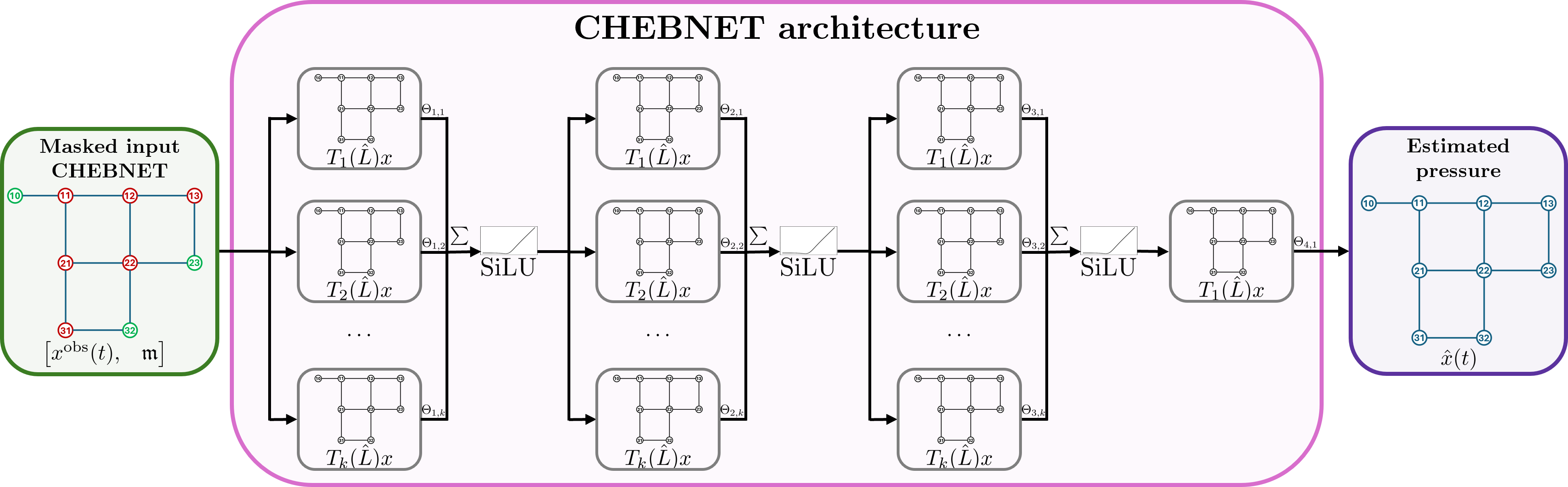}
    \caption{ChebNet architecture for pressure reconstruction and prediction in water distribution networks (WDNs).}
    \label{fig:ChebNet}
\end{figure*}
We now introduce the ChebNet architecture, which serves as the reconstruction–prediction model throughout this paper. 
ChebNet is a spectral graph neural network that applies Chebyshev polynomial filters to a scaled Laplacian of the WDN topology, enabling it to learn how pressures propagate across the junction graph from sparse measurements.
The graph convolution used in ChebNet begins with the normalized Laplacian based on the Laplacian defined in \eqref{eq: Laplacian}
\begin{equation*}
L_{\mathrm{norm}} = I - D^{-1/2} A D^{-1/2}.
\end{equation*}
constructed from the \emph{junction-only} WDN adjacency matrix  $A $ and degree matrix  $D $.
Because the eigenvalues of  $L_{\mathrm{norm}} $ lie in  $[0,2] $, it is rescaled
\begin{equation}\label{eq: Lhat}
    \hat L = \frac{2}{\lambda_{\max}} L_{\mathrm{norm}} - I,
\end{equation}
so that its eigenvalues lie within  $[-1,1] $, the region where Chebyshev polynomials are numerically stable \citep{hajgato2021reconstructing}. 
With this scaled operator in place, the Chebyshev polynomials used in the convolution are defined recursively as
\begin{equation}\label{eq: Chebyshev}
\begin{aligned}
T_0(\hat L) &= I, \qquad T_1(\hat L) = \hat L,\\
T_k(\hat L) &= 2 \hat L\, T_{k-1}(\hat L) - T_{k-2}(\hat L), \qquad k \ge 2.    
\end{aligned}
\end{equation}
A ChebNet layer then evaluates the truncated expansion
\begin{equation}\label{eq: GNNLayers}
    g_{\Theta}(\hat L)x = \sum_{k=0}^{K} \Theta_k\, T_k(\hat L)x,
\end{equation}
where  $K $ is the filter order and  $\Theta_k \in \mathbb{R}^{F_{\mathrm{in}} \times F_{\mathrm{out}}} $ are the learnable weights. This construction lets the network combine information across multiple hops in the junction graph, in a way that is directly governed by the WDN topology.

With these convolutional layers in place, we can train ChebNet to reconstruct nodal pressures from sparse measurements as follows. 
Let  $x(t) \in \mathbb{R}^n $ denote the pressure vector at time  $t $, and let  $\mask \in \{0,1\}^n $ indicate the sensor locations. The observed signal is
\begin{equation}\label{eq: sparseX}
    x^{\mathrm{obs}}(t) = \mask \odot x(t),
\end{equation}
where  $\odot $ denotes the Hadamard product.
From these sparse inputs, the reconstructor learns the mapping
\begin{equation}\label{eq ReconstProb}
    \hat{x}_r(t) = f_{\Theta}(x^{\mathrm{obs}}(t), \mask),
\end{equation}
which reconstructs the full pressure vector at time  $t $.
To train this model, we use pressure trajectories
\begin{equation*}
x_T=\{x(0),x(1),\ldots,x(T)\} \in \mathbb{R}^N,
\end{equation*}
generated using a hydraulic simulator such as EPANET.
Applying the mask \eqref{eq: sparseX} yields the inputs  $x^{\mathrm{obs}}(t) $, and the parameters of the reconstructor are optimized by minimizing
\begin{equation}\label{eq: TrainingRec}
    \mathcal{L}_{\mathrm{rec}}(\Theta)
= \frac{1}{T}\sum_{t=1}^{T}
\left\| f_{\Theta}(x^{\mathrm{obs}}(t), \mask) - x(t) \right\|_2^2.
\end{equation}
This model acts as a \emph{static} estimator, it reconstructs the pressure at time  $t $ using only the measurements available at that same instant.

To capture temporal behaviour instead of static, we extend this architecture to a \emph{predictor} that estimates  $x(t+1) $ from a window of past measurements. 
To formalize this, for a window length  $w \in \mathbb{N} $ we define
\begin{equation*}
\begin{aligned}
    x_{w}(t)
&= \big(x(t-w+1),\, x(t-w+2),\,\ldots,\, x(t)\big),\\
    x^{\mathrm{obs}}_w(t) &= \mask \odot x_w(t).
\end{aligned}
\end{equation*}
The predictor learns the mapping
\begin{equation}\label{eq:PredMapping}
    \hat{x}_{p}(t+1) = f_{\phi}\!\left(x^{\mathrm{obs}}_{w}(t),\, \mask\right).
\end{equation} 
For training the predictor, let  $T_{\mathrm{tr}} $ denote the set of time indices used for one-step prediction, and let  $N_{\mathrm{tr}} $ be its cardinality. The predictor parameters are obtained by minimizing
\begin{equation}\label{eq:TrainingPred}
    \mathcal{L}_{\mathrm{pred}}(\phi)
    \!=\! \frac{1}{N_{\mathrm{tr}}}
      \!\sum_{t \in T_{\mathrm{tr}}}\!
      \left\| f_{\phi}\!\left(x^{\mathrm{obs}}_{w}(t),\mask\right)
      - x(t+1) \right\|_2^{2}\!.
\end{equation}

Having both models in place, leakage detection is based on the residuals of the reconstruction and the prediction. We define the nodal residual
\begin{equation*}
    r_n(t) = \hat{x}_{r}(t) - \hat{x}_{p}(t),
\end{equation*}
which remains small under leak-free conditions. 
Because leaks occur along pipes rather than at junctions, these nodal residuals are projected onto edges
\begin{equation*}
    r_e(t) = \big| r_{n_i}(t) - r_{n_j}(t) \big|.
\end{equation*}
Finally, to reduce the effect of noise and short-term fluctuations, we apply a rolling window of length $m_r$ to the absolute edge residuals, yielding the smoothed residuals
\begin{equation*}
    \bar{r}_e(t)
    = \frac{1}{m_r}\sum_{\tau=t-m_r+1}^{t} r_e(\tau).
\end{equation*}
To detect abnormal behavior, i.e., a leak, we construct a threshold based on the leak-free training data. Let $\mu_e$ and $\sigma_e$ denote the mean and standard deviation of $ \bar{r}_e(t)$ over the healthy validation set. A leak alarm is raised on edge $e$ when
\begin{equation*}
    \bar r_e(t) > \mu_e + \alpha_a \sigma_e,
\end{equation*}
where $\alpha_a > 0$ is a tuning parameter that balances sensitivity and false-alarm rate.

\subsection{Consistent Graph Construction for PageRank and ChebNet}
\label{Subsec: GraphConstruction}
To connect the PageRank-Centrality-based sensor placement with the ChebNet reconstructor–predictor framework, both must operate on the same underlying graph.

Following the assumptions in Section \ref{Sec: Preliminaries and Problem Statement} and the definition of $\hat L$ in \eqref{eq: Lhat}, pressure sensors are placed only at junctions, while tanks, reservoirs, and pumps act as boundary nodes. Extracting the junction–junction connectivity from the hydraulic model results in a symmetric adjacency matrix $A \in \{0,1\}^{n \times n}$, where $n$ is the number of junctions. This same matrix defines the transition matrix $M = AD^{-1}$ used in the PageRank iteration \eqref{eq: PageRank}, and through the normalized and scaled Laplacian, it also determines the ChebNet graph convolution.
For example, the junction-only adjacency matrix of the EPANET Net1 (Fig. \ref{fig:wdn_example}) is
\begin{equation}\label{eq:ANet1_adj}
A =
\begin{bmatrix}
0 & 1 & 0 & 0 & 0 & 0 & 0 & 0 & 0\\
1 & 0 & 1 & 0 & 1 & 0 & 0 & 0 & 0\\
0 & 1 & 0 & 1 & 0 & 1 & 0 & 0 & 0\\
0 & 0 & 1 & 0 & 0 & 0 & 1 & 0 & 0\\
0 & 1 & 0 & 0 & 0 & 1 & 0 & 1 & 0\\
0 & 0 & 1 & 0 & 1 & 0 & 1 & 0 & 1\\
0 & 0 & 0 & 1 & 0 & 1 & 0 & 0 & 0\\
0 & 0 & 0 & 0 & 1 & 0 & 0 & 0 & 1\\
0 & 0 & 0 & 0 & 0 & 1 & 0 & 1 & 0
\end{bmatrix}.
\end{equation}
Using this unified adjacency matrix ensures full consistency: the PageRank-Centrality-based sensor placement selects junctions based on the same connectivity that determines how ChebNet propagates information through its spectral filters. PageRank uses $A$ to drive the random-walk dynamics, and ChebNet builds its Laplacian filters directly from $A$. As a result, reconstruction, prediction, and leakage detection all operate on a graph that is structurally aligned with the sensor placement strategy, ensuring that sensor placement and ChebNet filtering are fully aligned in terms of the underlying graph structure.

\section{Simulation results}\label{Sec: Simulation results}
In this section, we compare the PageRank-Centrality-based sensor placement with an arbitrary placement of equal size on the EPANET Net1 network. For clarity, the analysis is structured into three parts: (i) the sensor placement and training setup, (ii) validation and testing of the reconstructor and predictor, and (iii) leakage detection under a simulated fault.

\subsection{Sensor placement and training}

To obtain the PageRank-centrality-based sensor placement in Algorithm \ref{Alg:PageRankSensorPlacement}, we compute the PageRank scores of the junction-only adjacency matrix in \eqref{eq:ANet1_adj}, using a damping factor $\alpha_p = 0.85$. Table \ref{tab:PageRank} shows the resulting ranking, ordered from lowest to highest PageRank value. Selecting the three lowest-ranked junctions yields the sensor set $\color{PageRank}\{\mathbf{10},\,\mathbf{23},\,\mathbf{32}\}$. For comparison, we include an arbitrary placement of the same size, chosen to have a similar spatial distribution without overlapping sensor locations, namely $\color{Arbitrary}\{\mathbf{13},\,\mathbf{22},\,\mathbf{31}\}$. 
Note that the node identifiers in Table \ref{tab:PageRank} correspond to the EPANET junction labels (as in Fig. \ref{fig:wdn_example}), not to the row or column indices of the adjacency matrix.

\begin{table}[H]
\centering
\setlength{\tabcolsep}{2pt}
\caption{PageRank values for Net1 in Fig. \ref{fig:wdn_example}, computed using $A$ in \eqref{eq:ANet1_adj} with $\alpha_p = 0.85$.}
\label{tab:PageRank}
\begin{tabular}{c|ccccccccc}
\hline
Node  
& \textcolor{PageRank}{\textbf{10}}
& \textcolor{PageRank}{\textbf{23}}
& \textcolor{PageRank}{\textbf{32}}
& \textcolor{Arbitrary}{\textbf{13}}
& \textcolor{Arbitrary}{\textbf{31}}
& \textbf{21}
& \textbf{12}
& \textbf{11}
& \textcolor{Arbitrary}{\textbf{22}} \\
\hline
$p$   
& \textcolor{PageRank}{0.056}
& \textcolor{PageRank}{0.092}
& \textcolor{PageRank}{0.092}
& \textcolor{Arbitrary}{0.093}
& \textcolor{Arbitrary}{0.093}
& 0.132
& 0.132
& 0.139
& \textcolor{Arbitrary}{0.1699} \\
\hline
\end{tabular}
\end{table}\vspace{-0.2cm}

For each sensor configuration, we train a ChebNet reconstructor and predictor, resulting in four GNNs in total. Using EPANET, we generate a $108$-hour dataset with a 1-minute sampling interval ($6480$ samples), of which the first $60$ hours are used for training, hours $60$–$84$ for validation, and the final $24$ hours for testing. The predictor uses a temporal window of $w=120$ minutes.
All GNNs use a four-layer ChebNet architecture with Chebyshev orders $K=[48,\,24,\,4,\,1]$ and channel widths $F=[24,\,12,\,6]$, representing a scaled-down but structurally consistent version of the design used for the L-Town network in \citep{gardharsson2022graph}.

\begin{figure}[t]
\centering
\includegraphics[width=0.8\linewidth]{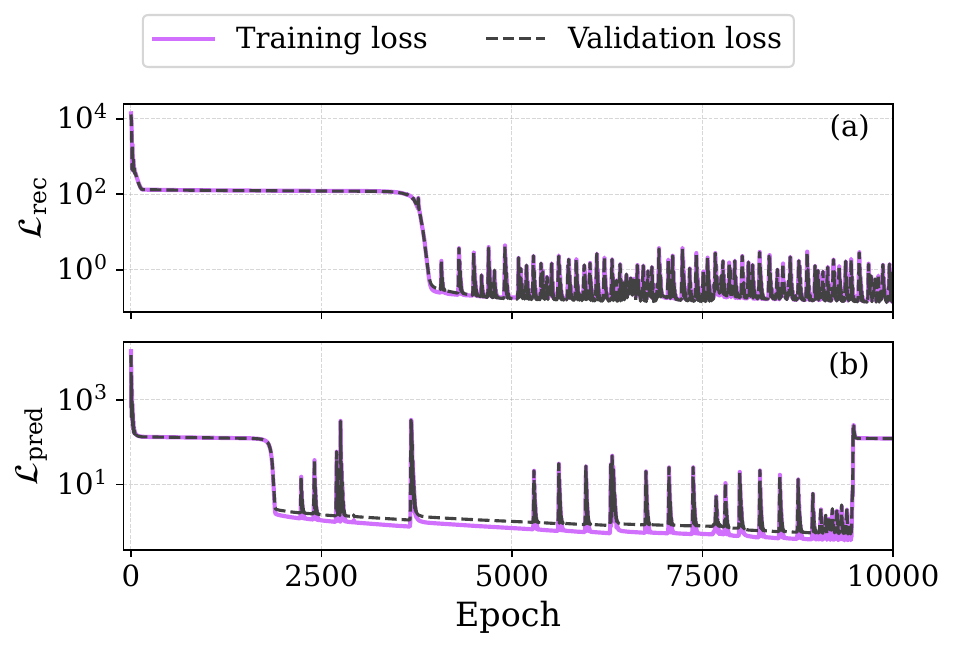}
\caption{Training and validation loss on a logarithmic scale for the reconstructor (a) and predictor (b), using PageRank-Centrality-based sensor placement.}
\label{fig:training_curves}
\end{figure}

\begin{table}[h!]
\centering
\setlength{\tabcolsep}{2pt}   
\caption{Training and validation mean squared errors (MSE) for the four GNNs on Net1.}
\label{tab:results_training}
\begin{tabular}{l|cc}
\hline
GNN & Training MSE & Validation MSE \\
\hline
Reconstructor (PageRank) & 0.1452 & \textbf{0.1358} \\
Reconstructor (Arbitrary)      & 3.910  & 4.407  \\
Predictor (PageRank)     & 0.4092 & \textbf{0.6831} \\
Predictor (Arbitrary)          & 2.848  & 3.582  \\
\hline
\end{tabular}
\end{table}
Table \ref{tab:results_training} summarizes the final training and validation errors for all four GNNs. The reconstructor and predictor trained with the PageRank-Centrality-based sensor placement achieve consistently lower validation errors than those trained with arbitrary placement, indicating that the PageRank-Centrality-based sensor configuration provides more informative measurements.
Additionally, for both sensor configurations, the predictor results in higher errors than the reconstructor. This behavior is also visible in the training curves in Fig. \ref{fig:training_curves}. The reconstructor converges quickly and stabilizes at a low loss with only minor fluctuations, whereas the predictor shows larger oscillations and a higher overall loss.

\begin{figure}[b]
    \centering
    \includegraphics[width=0.85\linewidth]{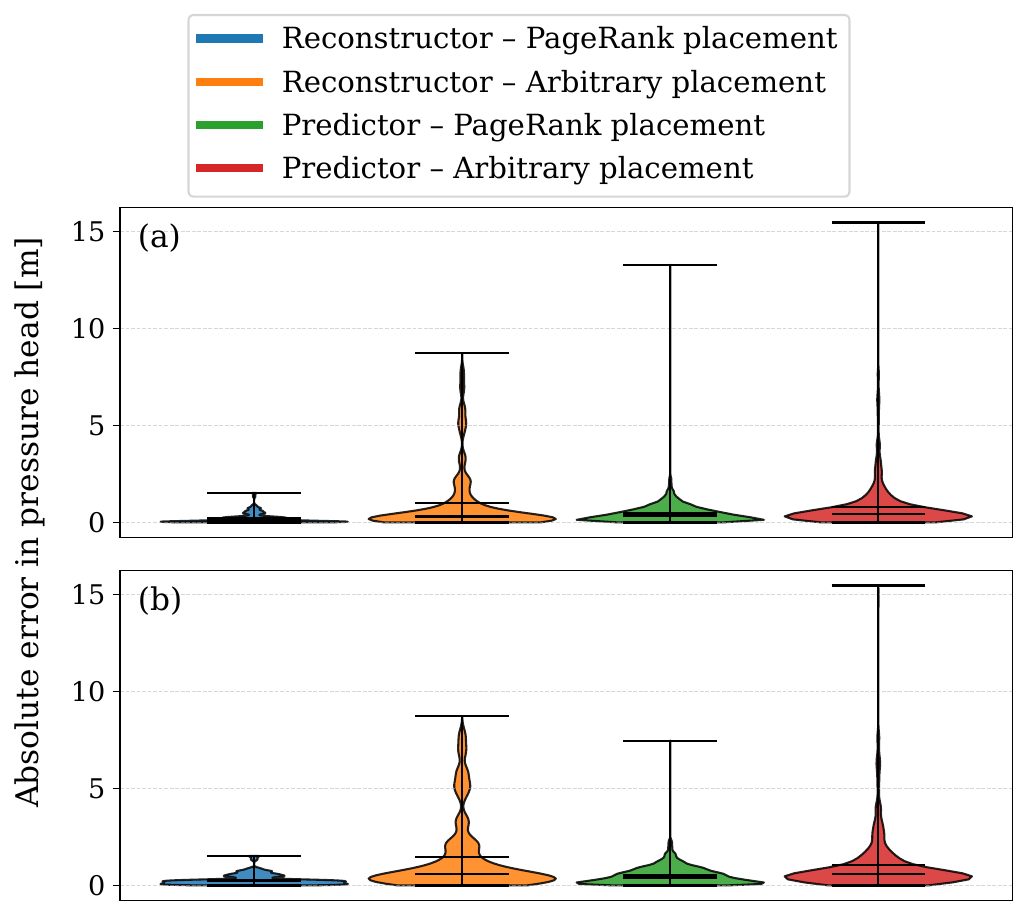}
    \caption{Residual distributions for the four GNNs on Net1, shown as violin plots over all junctions in (a) and over non-sensed junctions in (b). }
    \label{fig:violin}
\end{figure}

\subsection{Testing of the reconstructor and predictor}
\begin{figure}[t]
    \centering
    \includegraphics[width=0.9\linewidth]{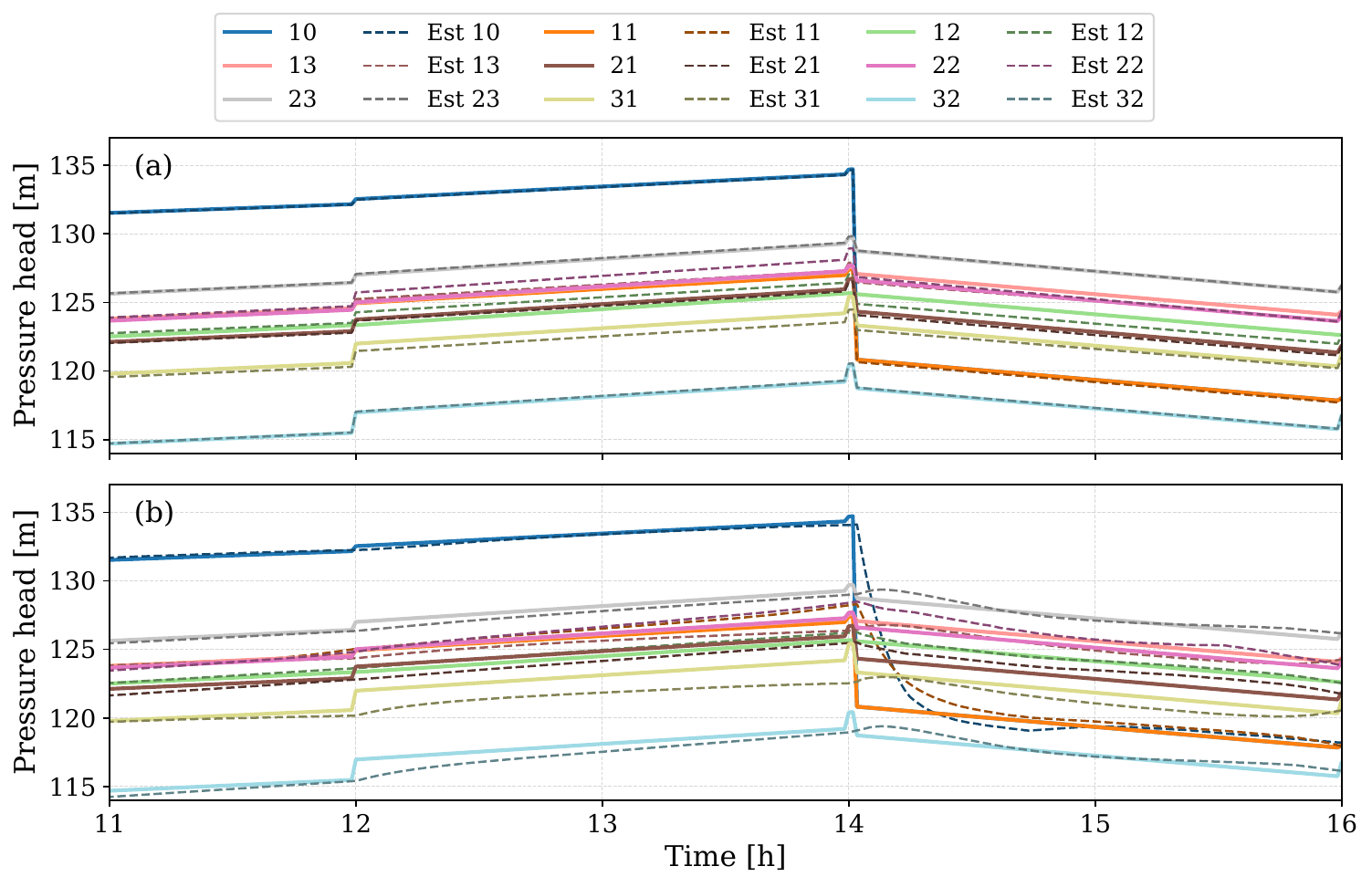}
    \caption{True vs. reconstructed pressures in (a) and true vs. one-step-ahead predictions in (b).}
    \label{fig:true_vs_est}
\end{figure}

To compare the reconstruction and prediction performance, we examine the residual distributions of all four GNNs by showing violin plots for all junctions and for the non-sensed junctions in Figure \ref{fig:violin}.
From Figure \ref{fig:violin}, we observe the same behaviour as in Table \ref{tab:results_training}, i.e., the PageRank-Centrality-based sensor placement yields narrower and more concentrated residuals, indicating higher reconstruction and prediction accuracy, whereas the arbitrary placement results in broader distributions. 
Additionally, we again see that the predictor consistently produces larger residuals than the reconstructor, since it must learn one-step-ahead dynamics instead of a static mapping. This makes the predictor more sensitive to abrupt changes in the pressure trajectory, particularly at the demand-driven discontinuities, which are visible in Fig. \ref{fig:true_vs_est}.

\subsection{Leakage simulation setup and detection}
Having established the four GNNs, we now compare both sensor configurations in a leakage scenario. To this end, Net1 is simulated for 30 hours with a 1-minute resolution, and a single leak is introduced at hour 26 by activating the emitter coefficient at junction 21 (Fig. \ref{fig:wdn_example}). This junction is not sensed in both configurations but lies adjacent to a junction equipped with either a PageRank-Centrality-based sensor or an arbitrary sensor. The leak magnitude is chosen such that the resulting head drop exceeds typical demand-driven discontinuities (about 1.5 m), ensuring 
that the event is distinguishable from normal variations.

\begin{figure}[b]
    \centering
    \includegraphics[width=0.99\linewidth]{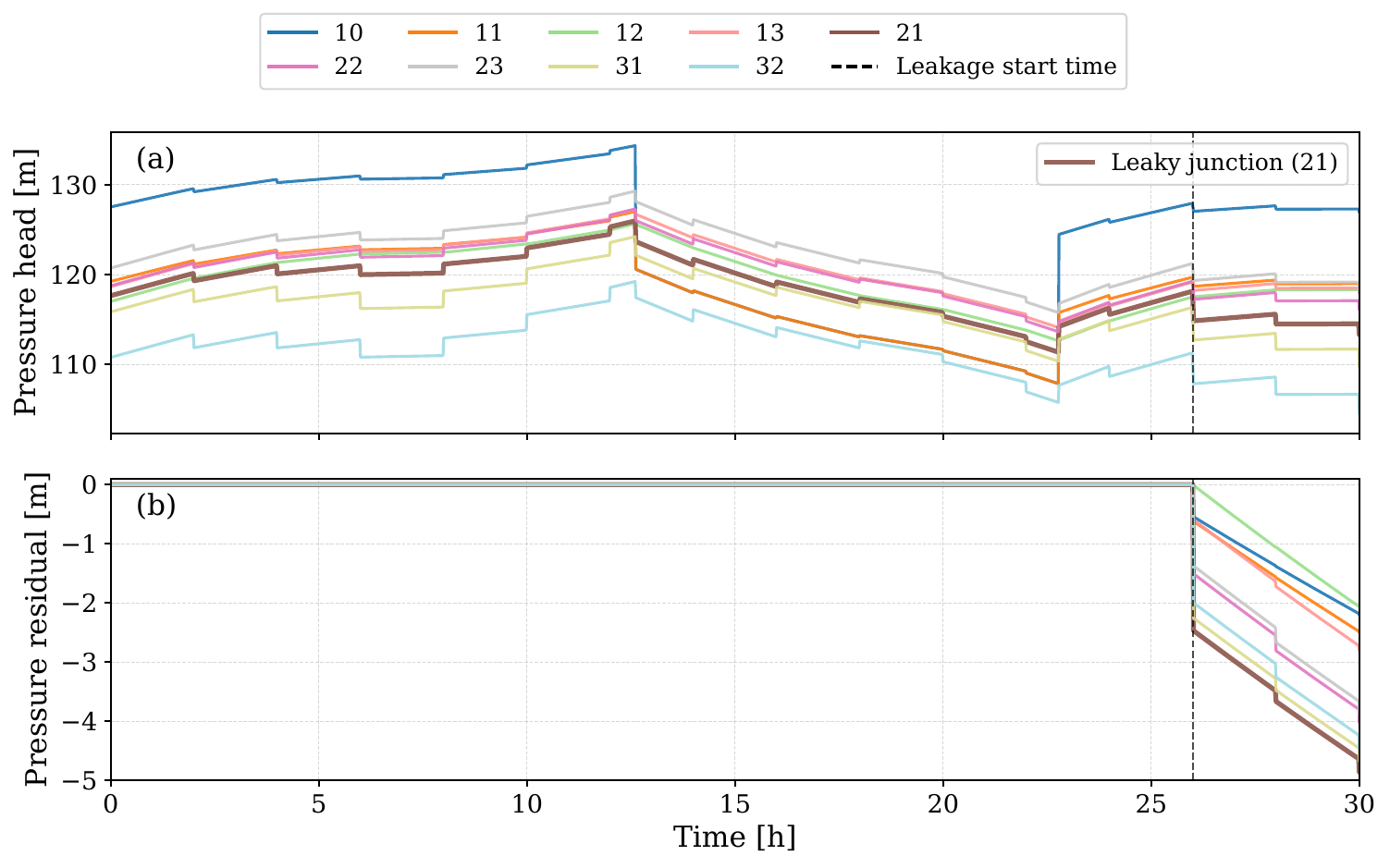}
    \caption{Leakage simulation in Net1: (a) pressure trajectories with the leaky junction highlighted, (b) pressure differences relative to leak-free conditions.}
    \label{fig:leaky_pressures}
\end{figure}

\begin{figure}[t]
    \centering
    \includegraphics[width=0.99\linewidth]{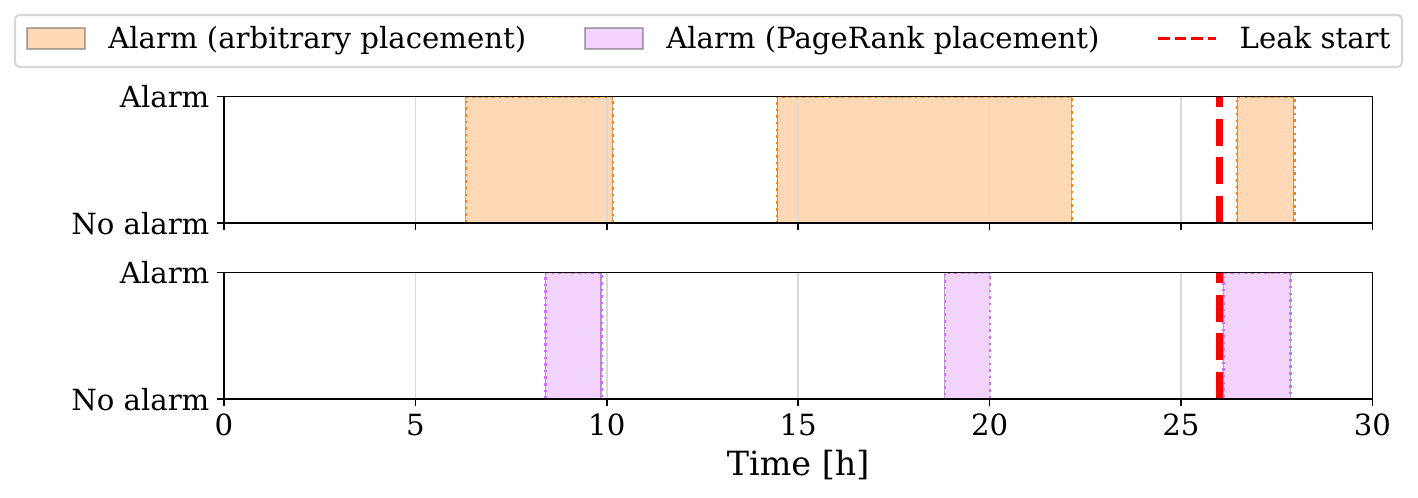}
    \caption{Comparison leak alarms for PageRank-Centrality-based and arbitrary sensor placements on Net1.}
    \label{fig:edge_fault_comparison}
\end{figure}

Figure \ref{fig:leaky_pressures} shows the resulting pressure trajectories and the corresponding pressure deviations, with the leaky junction highlighted. For both sensor configurations, we apply the same residual-based detection criterion introduced in Section \ref{Subsec: ChebNet Architecture for Pressure Reconstruction and Prediction} using $m_r=150$ and $\alpha_a=6$. The leakage alarm timelines are shown in Fig. \ref{fig:edge_fault_comparison}.
The red dashed line marks the leak start time at $26$h and the colored intervals indicate periods during which the residual-based detection triggers an alarm.

Before the leak starts, both configurations trigger false alarms due to the demand-driven pressure discontinuities that momentarily increase the prediction error. The PageRank-Centrality-based placement is noticeably less sensitive to these discontinuities, its alarms are shorter, and the detector returns more quickly to the no-alarm state. In contrast, the arbitrary placement reacts more strongly to the same discontinuities, resulting in longer false-alarm periods. 
Once the leak begins, the PageRank-Centrality-based placement raises an alarm immediately and maintains it for a longer interval than any of its false alarms. The arbitrary placement responds only after roughly 30 minutes and produces a shorter alarm, even though its false-alarm periods are longer. This behaviour reflects the improved reconstruction–prediction accuracy with the PageRank-Centrality-based sensor placement.

\section{Conclusion and Future Work}\label{Sec: Conclusion and Future Work}
This paper investigated how PageRank-Centrality-based sensor placement influences the performance of a ChebNet architecture for pressure reconstruction, one-step-ahead prediction, and leakage detection in water distribution networks. Using the EPANET Net1 benchmark, we demonstrated that selecting sensors based on PageRank Centrality yields more informative measurement locations, resulting in lower reconstruction and prediction errors and significantly reduced false-alarm time compared to an arbitrary sensor placement. These improvements were observed both in normal operation and under a leakage scenario. Additionally, the predictor consistently showed larger residuals than the reconstructor, which could be explained by the abrupt pressure discontinuities driven by the demand pattern.

For future work, several challenges could be explored. First, incorporating smoother and more realistic demand patterns may help reduce the discontinuity-driven prediction errors observed in the current setup. In practice, such sharp pressure steps rarely occur except during pipe bursts, normal demand variations produce much gentler changes. Second, further improvements to the architecture could focus on handling these sharp transients more effectively. Third, a robust method for distinguishing demand-driven pressure drops from leak-induced deviations remains an important open question. Finally, applying the PageRank-Centrality-based sensor placement to larger and more complex networks would allow for the assessment of scalability and generalization.

\bibliography{ifacconf}
\end{document}